%% file: main.tex
\documentclass{sig-alternate-05-2015}

\usepackage{times}
\usepackage{amsmath,amssymb}
\usepackage{latexsym}
\usepackage{graphicx}
\usepackage{comment}
\usepackage{hhline}
\usepackage{multirow}
\usepackage{dsfont}
\usepackage{scalerel}
\usepackage{balance}
\renewcommand{\vec}[1]{\mathbf{#1}}
\usepackage[pdfstartview=FitH]{hyperref} 
\usepackage[open,openlevel=2,numbered]{bookmark}

\begin{document}
%

\title{Deep Memory Networks for Attitude Identification}
%
%
%
%
%

\numberofauthors{3} 
%
\author{
%
%
\alignauthor
Cheng Li\\
       \affaddr{School of Information}\\
       \affaddr{University of Michigan\\ Ann Arbor}\\
       \email{lichengz@umich.edu}
\alignauthor
Xiaoxiao Guo\\
       \affaddr{Computer Science and Eng.}\\
       \affaddr{University of Michigan\\ Ann Arbor}\\
       \email{guoxiao@umich.edu}
\alignauthor 
Qiaozhu Mei\\
       \affaddr{School of Information}\\
       \affaddr{University of Michigan\\ Ann Arbor}\\
       \email{qmei@umich.edu}
}

\CopyrightYear{2017} 
\setcopyright{acmcopyright}
\conferenceinfo{WSDM 2017,}{February 06-10, 2017, Cambridge, United Kingdom}
\isbn{978-1-4503-4675-7/17/02}\acmPrice{\$15.00}
\doi{http://dx.doi.org/10.1145/3018661.3018714}
\maketitle

\clubpenalty=10000
\widowpenalty = 10000 

\input{tex/abstract}

%
%

\input{tex/Introduction}

\input{tex/Related_work}
\input{tex/ProposedApproach}
\input{tex/Experiment}
\input{tex/Experiment_result}

\input{tex/Conclusion}

%
\bibliographystyle{abbrv}
\small
\balance 
\bibliography{attitude_analysis_bib}  
%
%
\end{document}

%% file: tex/abstract.tex
\begin{abstract}
We consider the task of identifying attitudes towards a given set of entities from text. Conventionally, this task is decomposed into two separate subtasks: target detection that identifies whether each entity is mentioned in the text, either explicitly or implicitly, and polarity classification that classifies the exact sentiment towards an identified entity (the target) into positive, negative, or neutral.

Instead, we show that attitude identification can be solved with an end-to-end machine learning architecture, in which the two subtasks are interleaved by a deep memory network. In this way, signals produced in target detection provide clues for polarity classification, and reversely, the predicted polarity provides feedback to the identification of targets. Moreover, the treatments for the set of targets also influence each other -- the learned representations may share the same semantics for some targets but vary for others. The proposed deep memory network, the AttNet, outperforms methods that do not consider the interactions between the subtasks or those among the targets, including conventional machine learning methods and the state-of-the-art deep learning models. 

\end{abstract}

%% file: tex/Introduction.tex
\section{Introduction}
In many scenarios, it is critical to identify people's attitudes \footnote{ ``\textit{The way you think and feel about someone or something},'' as defined by Merriam-Webster. \url{http://www.merriam-webster.com/dictionary/attitude} } towards a set of entities. Examples include companies who want to know customers' opinions about their products, governments who are concerned with public reactions about policy changes, and financial analysts who identify daily news that could potentially influence the prices of securities. In a more general case, attitudes towards all entities in a knowledge base may be tracked over time for various in-depth analyses.

Different from a \textit{sentiment} which might not have a target (e.g., ``I feel happy'') or an \textit{opinion} which might not have a polarity (e.g., ``we should do more exercise''), an \textit{attitude} can be roughly considered as a sentiment polarity towards a particular entity (e.g., ``WSDM is a great conference''). 
Therefore, the task of attitude identification has been conventionally decomposed into two separate subtasks: target detection that identifies whether an entity is mentioned in the text, either explicitly or implicitly, and polarity classification that classifies the exact sentiment towards the identified target, usually into three categories: positive, negative, and neutral. 

Solving the two subtasks back-to-back is by no means unreasonable, but it may not be optimal. Specifically, intrinsic interactions between the two subtasks may be neglected in such a modularized pipeline. Indeed, signals identified in the first subtask -- both the words that refer to the target and the positions of these words, could provide useful information for the polarity of sentiments. For example, the identified target in the sentence ``\textit{this \underline{Tiramisu cake} is \underline{~~~}}'' indicates a high probability that a sentimental word would appear in the blank and is highly likely to be related to flavor or price. On the other hand, sentimental expressions identified in the second subtask and their positions could in turn provide feedback to the first task and signal the existence of a target. For example, the positive sentiment in ``\textit{the new \underline{Keynote} is user friendly} '' provides good evidence that ``\textit{Keynote}'' is a software (the target) instead of a speech (not the target). In addition, models learned for certain targets and their sentiments may share some important dimensions with each other while differ on other dimensions. For example, two targets \textit{food} and \textit{service} may share many sentimental expressions, but the sentence ``\textit{we have been waiting for food for one hour}'' is clearly about the service instead of the food. Failure to utilize these interactions (both between tasks and among targets) may compromise the performance of both subtasks. 

Recent developments of deep learning has provided the opportunity of a better alternative to modularized pipelines, in which machine learning and natural language processing tasks can be solved in an \textit{end-to-end} manner. With a carefully designed multi-layer neural network, learning errors backpropagate from upper layers to lower layers, which enables deep interactions between the learning of multi-grained representations of the data or multiple subtasks. Indeed, deep learning has recently been applied to target-specific sentiment analysis (mostly the second subtask of attitude identification) and achieved promising performance, where a given target is assumed to have appeared exactly once in a piece of text and the task is to determine the polarity of this text~\cite{vo2015target, zhang2016gated, tang2016aspect}. A deep network structure learns the dependency between the words in the context and the target word. 

In another related topic known as multi-aspect sentiment analysis, where the goal is to learn the fine-grained sentiments on different aspects of a target, some methods have attempted to model aspects and sentiments jointly. Aspects are often assumed to be mentioned explicitly in text, so that the related entities can be extracted through supervised sequence labeling methods~\cite{marcheggiani2014hierarchical,li2010structure,zirn2011fine}; aspects mentioned implicitly can be extracted as fuzzy representations through unsupervised methods such as topic models~\cite{mei2007topic,wang2011latent,sauper2013automatic}. While unsupervised methods suffer from low accuracy, it is usually difficult for supervised methods, like support vector machines (SVMs)~\cite{kobayashi2006opinion}, to interleave aspect extraction and sentiment classification. 


In this paper, we show that the accuracy of attitude identification can be significantly improved through effectively modeling the interactions between subtasks and among targets. The problem can be solved with an end-to-end machine learning architecture, where the two subtasks are interleaved by a deep memory network. The proposed model, called the AttNet, also allows different targets to interact with each other, by sharing a common semantic space and simultaneously keep their own space, making it possible for all targets to be learned in a unified model. The proposed deep memory network outperforms models that do not consider the subtask or target interactions, including conventional supervised learning methods and state-of-the-art deep learning models.

The rest of the paper is organized as follows. Section~\ref{sec:related} summarizes the related literature. In Section~\ref{sec:method}, we describe how the deep neural network is designed to incorporate the interaction both between subtasks and among targets.   We present the design and the results of empirical experiments in Section~\ref{sec:exp_setup} and Section~\ref{sec:exp}, and then conclude the paper in Section~\ref{sec:conclusions}.

%% file: tex/Related_work.tex
\section{Related work}
\label{sec:related}

Sentiment analysis has been a very active area of research~\cite{pang2008opinion, popescu2011dancing}. While sentiment in general does not need to have a specific target, the notion of \textit{attitude} is usually concerned with a sentiment towards a target entity (someone or something). As one category of sentiment analysis, there is much existing work related to attitude identification, which generally takes place in three domains: multi-aspect sentiment analysis in product reviews, stance classification in online debates, and target-dependent sentiment classification in social media posts. Below we categorize existing work by the problem settings, e.g., whether the target is required to be explicitly mentioned.

\textbf{Explicitly tagged targets.} There has been a body of work that classifies the sentiment towards a particular target that is explicitly mentioned and tagged in text, mostly applied to social media text such as Tweets. Due to the short length of Tweets, many models assume that targets appear exactly once in every post. Jiang et al.~\cite{jiang2011target} developed seven rule-based target-dependent features, which are fed to an SVM classifier. Dong et al.~\cite{dong2014adaptive} proposed an adaptive recursive neural network that propagates sentiment signals from sentiment-baring words to specific targets on a dependence tree. Vo et al.~\cite{vo2015target} split a Tweet into a left context and a right context according to a given target, and used pre-trained word embeddings and neural pooling functions to extract features. Zhang et al.~\cite{zhang2016gated} extended this idea by using gated recursive neural networks. The paper most relevant to ours is Tang et al.~\cite{tang2016aspect}, which applied Memory Networks~\cite{sukhbaatar2015end} to the task of multi-aspect sentiment analysis. Aspects are given as inputs, assuming that they have already been annotated in the text. Their memory network beat all LSTM-based networks but did not outperform SVM with hand-crafted features.

Model structures for target-dependent sentiment classification heavily rely on the assumption that the target appears in the text \textit{explicitly}, and exactly \textit{once}. These models could degenerate when a target is implicitly mentioned or mentioned multiple times. Additionally, they do not consider the interactions between the subtasks (target detection and sentiment classification) or among the targets.

\textbf{Given target, one per instance.} In the problem of stance classification, the target, mentioned explicitly or implicitly, is given but not tagged in a piece of text. The task is only to classify the sentiment polarity towards that target. Most methods train a specific classifier for each target and report performance separately per target. Many researchers focus on the domain of online debates. They utilized various features based on n-grams, part of speech, syntactic rules, and dialogic relations between posts~\cite{walker2012your,hasan2013stance,faulkner2014automated, rajadesingan2014identifying}. The workshop SemEval-2016 presented a task on detecting stance from tweets~\cite{mohammad2016semeval}, where an additional category is added for the given target, indicating the absence of sentiment towards the target. Mohammad et al.~\cite{mohammad2016stance} beat all teams by building an SVM classifier for each target.

As stance classification deals with only one given target per instance, it fails to consider the interaction between target detection and sentiment classification. Furthermore, the interplay among targets is ignored by training a separate model per target.

\textbf{Explicit targets, not tagged.} In the domain of product reviews, a specific aspect of a product could be considered as a target of attitudes. When the targets appear in a review but are not explicitly tagged, they need to be extracted first. Most work focuses on extracting explicitly mentioned aspects. Hu et al.~\cite{hu2004mining} extracted product aspects via association mining, and expanded seed opinion terms by using synonyms and antonyms in WordNet. When supervised learning approaches are taken, both tasks of aspect extraction and polarity classification can be cast as a binary classification problem~\cite{kobayashi2006opinion}, or as a sequence labeling task and solved using sequence learning models such as conditional random fields (CRFs)~\cite{marcheggiani2014hierarchical,li2010structure} or hidden Markov models (HMMs)~\cite{zirn2011fine}.

\textbf{Implicit targets.} There are studies that attempt to address the situation when aspects could be implicitly mentioned. Unsupervised learning approaches like topic modeling treat aspects as topics, so that topics and sentiment polarity can be jointly modeled~\cite{mei2007topic,wang2011latent,sauper2013automatic}. The workshop of SemEval-2015 announced a task of aspect based sentiment analysis~\cite{pontiki2015semeval}, which separates aspect identification and polarity classification into two subtasks. For aspect identification, top teams cast aspect category extraction as a multi-class classification problem with features based on n-grams, parse trees, and word clusters.

Although aspect identification and polarity classification are modeled jointly here, it is hard to train unsupervised methods in an end-to-end way and directly optimize the task performance. 

\textbf{Deep learning for sentiment analysis.} In the general domain of sentiment analysis, there has been an increasing amount of attention on deep learning approaches. In particular, Bespalov et al.~\cite{bespalov2011sentiment} used Latent Semantic Analysis to initialize the word embedding, representing each document as the linear combination of n-gram vectors. Glorot et al.~\cite{glorot2011domain} applied Denoising Autoencoders for domain adaptation in sentiment classification. A set of models have been proposed to learn the compositionality of phrases based on the representation of children in the syntactic tree~\cite{socher2012semantic,socher2013recursive,hermann2013role}. These methods require parse trees as input for each document. However, parsing does not work well on user generated contents, e.g., tweets~\cite{gimpel2011part}. Liu et al.~\cite{liu2015fine} used recurrent neural networks to extract explicit aspects in reviews. 

Compared to the existing approaches, our work develops a novel deep learning architecture that emphasizes the interplay between target detection and polarity classification, and the interaction among multiple targets. These targets can be explicitly or implicitly mentioned in a piece of text and do not need to be tagged a priori. 

%% file: tex/ProposedApproach.tex
\section{AttNet for Attitude Identification}
\label{sec:method}
%
We propose an end-to-end neural network model to interleave the target detection task and the polarity classification task. 
The \textbf{target detection} task is to determine whether a specific target occurs in a given context either explicitly or implicitly. The \textbf{polarity classification} task is to decide the attitude of the given context towards the specific target if the target occurs in the context. 
Formally, a \textbf{target detection classifier} is a function mapping pairs of targets and contexts into binary labels, $($\textrm{context}, \textrm{target}$)$ $\rightarrow$ $\{\textrm{present}, \textrm{absent}\}$. A \textbf{polarity classifier} is a function mapping pairs of targets and contexts into three attitude labels, $($\textrm{context}, \textrm{target}$)$ $\rightarrow$ $\{\textrm{positive}, \textrm{negative},$ $ \textrm{neutral}\}$. 
For example, given a \textbf{context}, \textit{if everyone has guns, there would be just mess}, and a \textbf{target}, \textit{gun control}, the correct label is \textit{present} for the target detection task and \textit{positive} for polarity classification.  

Our model builds on the insight that the target detection task and the polarity classification task are deeply coupled in several ways. 
\begin{itemize}
\item
The polarity classification depends on the target detection because the polarity is meaningful only if the target occurs in the context. 
Reversely, the polarity classification task provides indirect supervision signals for the target detection task. For example, if the attitude label \textit{positive} is provided for a context-target pair, the target must have occurred in the context following the definition. Such indirect supervision signals are useful especially when the target only occurs in the context implicitly.
\item
The signal words in the target detection and the polarity classification task are usually position-related: the signal words to determine the polarity are usually the surrounding words of the signal words to detect the target. Moreover, when a context has multiple targets, the signal words usually cluster for different targets~\cite{hu2004mining, popescu2011dancing}.
\item
Different targets interact in both the target detection task and the polarity classification task. Intuitively, some context words could mean the same for many targets, while some context words could mean differently for different targets. 

\end{itemize}

Specifically, our model introduces several techniques building on the interaction between the target detection task and the polarity classification task accordingly.
\begin{itemize}
\item  The output of the target detection is concatenated as part of the input of the polarity classification task to allow polarity classification to be conditioned on target detection. Polarity classification labels are also used to train the target detection classifier by back-propagating the errors of the polarity classification to the target detection end-to-end. 
\item The attention of polarity classification over context words are preconditioned by the attention of target detection. The polarity classification task benefits from such precondition especially when there are multiple targets in the context.
\item  Target-specific projection matrices are introduced to allow some context words to have similar representations among targets and other context words to have distinct representations. These matrices are all learned in an end-to-end fashion.
\end{itemize}

We propose a deep memory network model, called the AttNet, which implements the above motivation and ideas. 
In the rest of this section, we give a brief introduction to the memory network model, followed by a description of a single layer version of the model. Then we extend the expressiveness and capability of the model by stacking multiple layers. 


\subsection {Background: Memory Networks}
As one of the recent developments of deep learning, memory networks~\cite{sukhbaatar2015end} have been successfully applied to language modeling, question answering, and aspect-level sentiment analysis~\cite{tang2016aspect}, which generates superior performance over alternative deep learning methods, e.g., LSTM. 

Given a context (or document, e.g., ``\textit{we have been waiting for food for one hour}'') and a target (e.g., \textit{service} ), a memory network layer converts the context into a vector representation by computing a weighted sum of context word vector representations. The weight is a score that measures the relevance between the context word and the target (e.g., a higher score between the words \textit{waiting} and \textit{service}). The vector representation of the context is then passed to a classifier for target detection or polarity classification. 
An attractive property is that all parameters, including the target embeddings, context word embeddings and scores, are end-to-end trainable without additional supervision signals.

AttNet improves the original memory network models for attitude identification by (1) interleaving the target detection and polarity classification subtasks and (2) introducing target-specific projection matrices in representation learning, without violating the end-to-end trainablity. 

\subsection {Single Layer AttNet}
\begin{figure*}[t]
\centering
\includegraphics[width=0.86\textwidth]{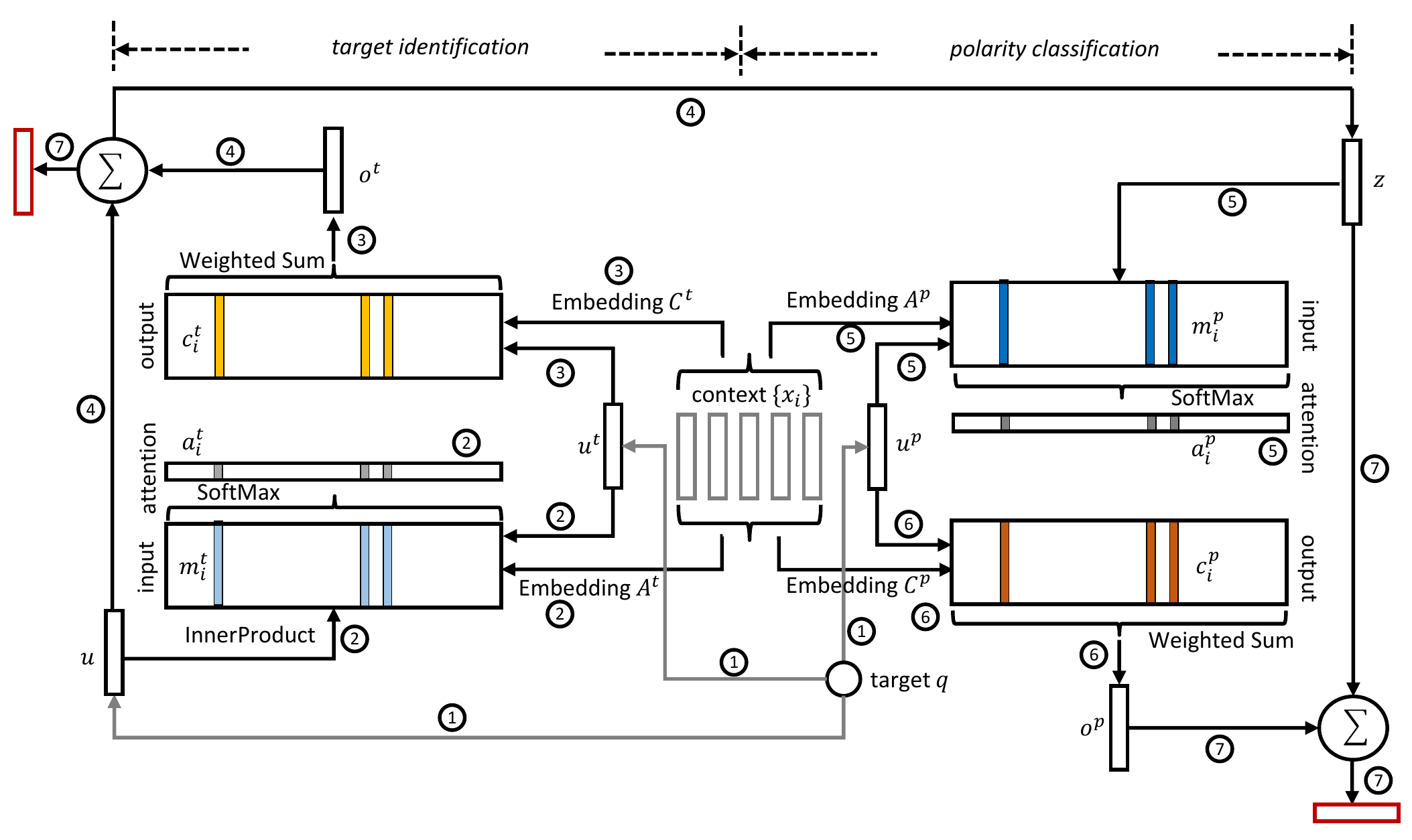}
\caption{A single layer version of AttNet. Key submodules are numbered and correspondingly detailed in the text. \label{fig:single}}
\end{figure*}

We begin by describing AttNet in the single layer case, shown in Figure~\ref{fig:single}. 
Hereafter for simplicity, we refer to the task of target detection as \textbf{TD}, and polarity classification as \textbf{PC}. 


\paragraph*{\bf (1) Target Embedding} Each query target is represented as a one-hot vector, $\vec{q} \in \mathds{R}^{N_{\textrm{target}}}$, where $N_{\textrm{target}}$ is the number of targets. All targets share a target embedding matrix $\vec{B} \in \mathds{R}^{d\times N_{\textrm{target}}}$, where $d$ is the embedding dimensionality. The matrix $\vec{B}$ converts a target into its embedding vector $\vec{u}=\vec{Bq}$, which is used as the input for the \textbf{TD} task.


\paragraph*{\bf (2) Input Representation and Attention for \textbf{TD}}  We compute match scores between the context (or document) and the target for content-based addressing. 
The context is first converted into a sequence of one-hot vectors, $\{\vec{x}_{i} \in \mathds{R}^{N_{\textrm{voc}}} \}$ , where $\vec{x}_{i}$ is the one-hot vector for the $i$-th word in the context and $N_{\textrm{voc}}$ is the number of words in the dictionary. 
The entire set of $\{\vec{x}_{i}\}$ are then embedded into a set of input representation vectors $\{\vec{m}^{t}_{i}\}$ by:
\[
\vec{m}_{i}^{t} = \vec{V}^{t}_{q}\vec{A}^{t} \vec{x}_{i},
\]
where $\vec{A}^{t} \in \mathds{R}^{d\times N_{\textrm{voc}}}$ is the word embedding matrix shared across targets, superscript $t$ stands for the \textbf{TD} task, and $\vec{V}^{t}_{q} \in \mathds{R}^{d\times d}$ is a target-specific projection matrix for target $q$, which allows context words $\vec{x}_{i}$ to share some semantic dimensions for some targets while vary for others. 

In the embedding space, we compute the match scores between the target input representation $\vec{u}$ and each context word representation $\vec{m}_{i}^{t}$ by taking the inner product followed by a softmax, $a^{t}_{i} = \textrm{SoftMax}(\vec{u}^{\intercal} \vec{m}_{i}^{t}) $, 
where $\textrm{SoftMax}(w_{i}) = \exp(w_{i})  / \sum_{j} \exp(w_{j})$. In this way, $\vec{a}^{t}$ is a soft attention (or probability) vector defined over the context words. 

\paragraph*{\bf (3) Output Representation for \textbf{TD}} A different embedding matrix, $\vec{C}^{t} \in \mathds{R}^{d\times N_{\textrm{voc}}}$, is introduced for flexibility in computing the output representation of context words by: 
\[
\vec{c}_{i}^{t} = \vec{V}^{t}_{q} \vec{C}^{t} \vec{x}_{i}
\]
The response output vector $\vec{o}^{t}$ is then a sum over the outputs $\vec{c}_{i}^{t}$, weighted by the attention vector from the input: $ \vec{o}^{t} = \sum_{i} a^{t}_{i} \vec{c}_{i}^{t}$. 

\paragraph*{\bf (4) Interleaving \textbf{TD} and \textbf{PC}} In the single layer case, the sum of the output vector $\vec{o}^{t}$ and the target query embedding $\vec{u}$ is then passed to the \textbf{PC} task, $ \vec{z} = \vec{o}^{t} + \vec{u} $.

\paragraph*{\bf (5) Input Representation and Attention for \textbf{PC}} Similar to the \textbf{TD} task, we convert the entire set of $\{\vec{x}_{i}\}$ into input representation vectors $\{\vec{m}^{p}_{i}\}$ by: 
\[
\vec{m}_{i}^{p} = \vec{V}^{p}_{q} \vec{A}^{p}\vec{x}_{i},
\]
where $\vec{A}^{p} \in \mathds{R}^{d\times N_{\textrm{voc}}}$ is the input embedding matrix for \textbf{PC}. We use separate embedding matrices $\vec{A}^{t}$ and $\vec{A}^{p}$ for \textbf{TD} and \textbf{PC}, as the words could have different semantics in the two tasks. For similar reasons, we use different projection matrices $\vec{V}^{t}_{q}$ and $\vec{V}^{p}_{q}$ for the two tasks.

Given the polarity input representation $\{\vec{m}^{p}_{i}\}$, we also compute the soft attention over the context words for polarity identification, $ a^{p}_{i} = \textrm{SoftMax}(\vec{z}^{\intercal} \vec{m}_{i}^{p})$. 

\paragraph*{\bf (6) Output Representation for \textbf{PC}}
There is also one corresponding output vector $\vec{c}^{p}_{i}$ in \textbf{PC} for each $\vec{x}_{i}$:  
\[
\vec{c}_{i}^{p} = \vec{V}^{p}_{q} \vec{C}^{p} \vec{x}_{i},
\]
where $\vec{C}^{p} \in \mathds{R}^{d\times N_{\textrm{voc}}}$ is the polarity output embedding matrix. It has been observed that sentiment-baring words are often close to the target~\cite{hu2004mining, popescu2011dancing}. Based on this observation, attentions, or positions of important words that identify the target in the first module, could provide prior knowledge to learn the attention of the second module. Therefore we compute the final attention vector as a function of original attentions of both tasks:
\begin{equation}
\vec{b}^p = (1-\lambda)\vec{a}^p + \lambda f(\vec{a}^t),
\label{equ:prior_attention}
\end{equation}
where $0 \le \lambda < 1$ controls the importance of the second term, and $f$ is a moving average function which shifts attentions from words of high values to their surrounding neighbors. The output vector is $\vec{o}^{p} = \sum_{i} b^{p}_{i} \vec{c}_{i}^{p}$.

\paragraph*{\bf (7) Prediction for \textbf{TD} and \textbf{PC}}
To predict whether a target presents, the sum of the output vector of target classification $\vec{o}^{t}$ and the target query vector $\vec{u}$ is passed through a weight matrix $\vec{W}^{t} \in \mathds{R}^{2\times d}$ (2 is the number of classes: \textit{present}, \textit{absent}) and a softmax operator to produce the predicted label, a vector of class probabilities: $\vec{y}^{t} = \textrm{SoftMax}(\vec{W}^{t}(\vec{o}^{t}+\vec{u}))$.

Similarly, the sum of the output vectors $\vec{o}^{p}$ of \textbf{PC} and its input vector $\vec{z}$ is then passed through a weight matrix $\vec{W}^{p} \in \mathds{R}^{3 \times d}$ and a softmax operator to produce the predicted attitude label vector, $\vec{y}^{p} = \textrm{SoftMax}(\vec{W}^{p}(\vec{o}^{p}+\vec{z}))$.

\subsection{Multiple Layer AttNet}
\begin{figure}[t]
\includegraphics[width=0.47\textwidth]{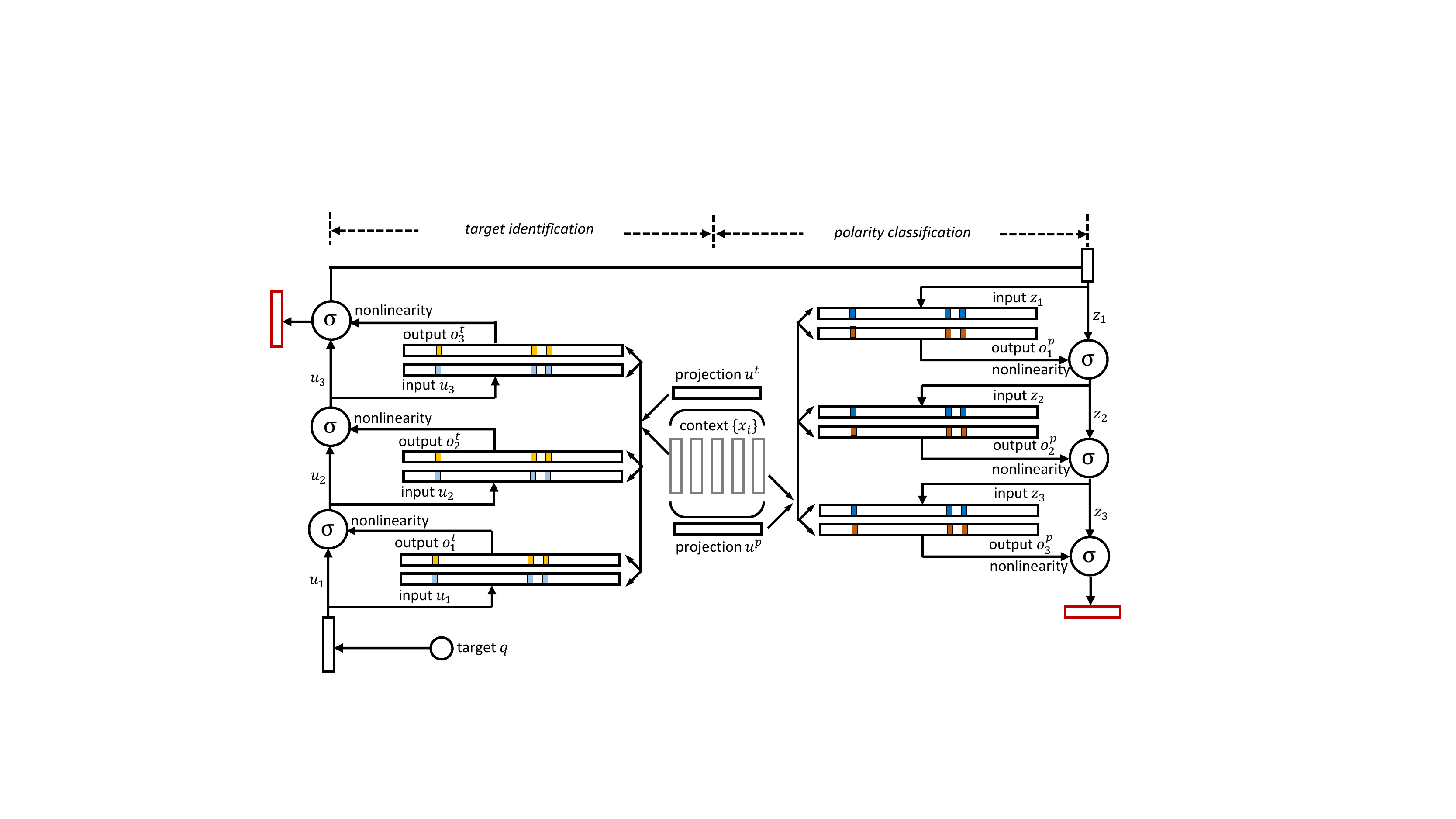}
\caption{A three layer version of our model. Both the \textbf{TD} and \textbf{PC} modules have three stacked layers. \label{fig:multiple}}
\end{figure}

We now extend our model to stacked multiple layers. Figure~\ref{fig:multiple} shows a three layer version of our model. The layers are stacked in the following way:
\paragraph*{\bf Functionality of Each Layer} For \textbf{TD}, the input to the $(k\scriptsize{+}1)$-th layer is the sum of the output $\vec{o}^{t}_{k}$ and the input $\vec{u}_{k}$ from the $k$-th layer, followed by a sigmoid nonlinearity: $\vec{u}_{k+1}=\sigma(\vec{H}^{t}\vec{u}_{k}+\vec{o}^{t}_{k})$, where $\sigma(x) = 1 / (1+\exp(x))$ is the sigmoid function and $\vec{H}^{t}$ is a learnable linear mapping matrix shared across layers. For the \textbf{PC} task, the input to the first layer is the transformed sum from the last layer of the \textbf{TD} module, $\vec{z}_{1} = \sigma(\vec{H}^{t}\vec{u}_{K_t}+\vec{o}^{t}_{K_t})$, where $K_t$ is the number of stacked layers in the \textbf{TD} task. Thus the prediction of polarity would depend on the output of the \textbf{TD} task and reversely the \textbf{TD} task would benefit from indirect supervision from the \textbf{PC} task by backward propagation of errors. Similarly for \textbf{PC}, the input to the $(k\scriptsize{+}1)$-th layer is the sum of the output $\vec{o}^{p}_{k}$ and the input $\vec{z}_{k}$ from the $k$-th layer, followed by a sigmoid nonlinearity: $\vec{z}_{k+1} = \sigma(\vec{H}^{p}\vec{z}_{k} + \vec{o}^{p}_{k})$.

\paragraph*{\bf Attention for \textbf{PC}} In the single layer case, the attention for \textbf{PC} is based on that of the \textbf{TD} module. When layers are stacked, all layers of the first module collectively identify important attention words to detect the target. Therefore we compute the averaged attention vector across all layers in the \textbf{TD} module $\bar{\vec{a}}^t = \frac{1}{K_t}\sum_{k=1}^{K_t}\vec{a}^t_k$. Accordingly for $k$-th layer of the \textbf{PC} module, the final attention vector is $\vec{b}^p_k = (1-\lambda)\vec{a}_k^p + \lambda f(\bar{\vec{a}}^t)$, and the output vector is $\vec{o}^{p}_k = \sum_{i} b^{p}_{k,i} \vec{c}_{k,i}^{p}$.

\paragraph*{\bf Tying Embedding and Projection Matrices} 
The embedding matrices and projection matrices are constrained to ease training and reduce the number of parameters~\cite{sukhbaatar2015end}. The embedding matrices and the projection matrices are shared for different layers. Specifically, using subscription $(k)$ denote the parameters in the $k$-th layer, for any layer $k$, we have $\vec{A}^{t (1)} \equiv \vec{A}^{t (k)}$, 
$\vec{C}^{t(1)} \equiv \vec{C}^{t(k)}$, $\vec{A}^{p(1)} \equiv \vec{A}^{p(k)}$, $\vec{C}^{p(1)} \equiv  \vec{C}^{p(k)}$, $\vec{V}^{t(1)}_{q} \equiv \vec{V}^{t(k)}_{q}$ and $\vec{V}^{p(1)}_{q} \equiv \vec{V}^{p(k)}_{q}$.

\paragraph*{\bf Predictions for \textbf{TD} and \textbf{PC}}
The prediction stage is similar to the single-layer case, with the prediction based on the output of the last layer $K_t$ (for \textbf{TD}) and $K_p$ (for \textbf{PC}). For the \textbf{TD} task, $\vec{y}^{t} = \textrm{SoftMax}(\vec{W}^{t}\sigma(\vec{H}^{t}\vec{u}_{K_t} + \vec{o}^{t}_{K_t}))$, 
while for \textbf{PC},  $\vec{y}^{p} = \textrm{SoftMax}(\vec{W}^{p}$ $\sigma(\vec{H}^{p}\vec{z}_{K_p}+\vec{o}^{p}_{K_p}))$.


\subsection{End-to-End Multi-Task Training}
We use cross entropy loss to train our model end-to-end given a set of training data $\{ct_{i}, q_{j}, g^{t}_{ij}, g^{p}_{ij}\}$, where $ct_{i}$ is the $i$-th context (or document), $q_{j}$ is the $j$-th target, $g^{t}_{ij}$ and $g^{p}_{ij}$ are the ground-truth labels for the \textbf{TD} and the \textbf{PC} tasks respectively. The training is to minimize the objective function:
\[
\mathcal{L} = - \sum_{i}\sum_{j} \Big( \log(\vec{y}^{t}_{ij}(g^{t}_{ij})) + \mathds{1}_{g_{ij}^{t}} \log(\vec{y}^{p}_{ij}(g^{p}_{ij})) \Big),
\] 
where $\vec{y}^{t}_{ij}$ is a vector of predicted probability for each class of \textbf{TD}, $\vec{y}^{t}_{ij}(s)$ selects the $s$-th element of $\vec{y}^{t}_{ij}$, $\mathds{1}_{g_{ij}^{t}}$ equals to 1 if $g_{ij}^{t}$ equals to class \textit{present} and 0 otherwise. Note that when a target is not mentioned in a given context, the polarity term plays no role in the objective because the value of $\mathds{1}_{g_{ij}^{t}}$ is zero.




%

%% file: tex/Experiment.tex
\section{Experiment Setup}
\label{sec:exp_setup}
In the experiments, we compare AttNet to conventional approaches and alternative deep learning approaches on three real world data sets, and we show the superior performance of our model. We also experiment with variants of AttNet as credit assignments for the key components in our model.
\subsection{Data Sets}
We examine AttNet on three domains that are related to attitude classification: online debates (Debates), multi-aspect sentiment analysis on product review (Reviews), and stance in tweets (Tweets). 

\textbf{Debates.} This data set is from the Internet Argument Corpus version 2\footnote{\url{https://nlds.soe.ucsc.edu/iac2}.}.  
The data set consists of political debates on three Internet forums\footnote{4forums(\url{http://www.4forums.com/political/}), \\  ConvinceMe(\url{http://www.convinceme.net/}) and \\ CreateDebate(\url{http://www.createdebate.com/})} .
In these forums, a person can initiate a debate by posting a topic and taking positions such as \textit{favor} vs. \textit{against}. Examples of topics are \textit{gun control}, \textit{death penalty} and \textit{abortion}. Other users participate in these debates by posting their arguments for one of the sides. 

\textbf{Tweets.} This data set comes from a task of the workshop SemEval-2016 on detecting stance from tweets~\cite{mohammad2016semeval}. Targets are mostly related to ideology, e.g., \textit{atheism} and \textit{feminist movement}\footnote{Since there are less than 10 tweets with neutral stance, we only consider \textit{positive} and \textit{negative} attitude by discarding these neutral tweets.}. 

\textbf{Review.} This data set includes reviews of restaurants and laptops from SemEval 2014~\cite{pontiki2014semeval} and 2015~\cite{pontiki2015semeval}, where subtasks of identifying aspects and classifying sentiments are provided. We merge two years' data to enlarge the data set, and only include aspects that are annotated in both years.

To guarantee enough training and test instances, for all the data sets we filter out targets mentioned in less than 100 documents. The original train-test split is used if provided, otherwise we randomly sample 10\% data into test set. We further randomly sample 10\% training data for validation. Text pre-processing includes stopword removal and tokenization by the CMU Twitter NLP tool~\cite{gimpel2011part}. The details of the data sets are shown in Table~\ref{tab:dataset}.

\begin{table}[h!]
\caption{Statistics of each data set.}
\small
\begin{center}
\setlength{\tabcolsep}{3.5pt}
\begin{tabular}{c|c|c|c|c|c|c}
\hline Data set & Set & \#docs & \#pos & \#neg & \#neutral & \#absent \\ 
\hline \multirow{3}{*}{Debates} & train & 24352 & 13891 & 10711 & 0 & 0  \\ \cline{2-7}
 & val & 2706 & 1530 & 1203 & 0 & 0 \\ \cline{2-7}
 & test & 3064 & 1740 & 1371 & 0 & 0 \\
\hline \multirow{3}{*}{Tweets} & train & 2614 & 682 & 1253 & 0 & 679 \\ \cline{2-7}
 & val & 291 & 71 & 142 & 0 & 78  \\ \cline{2-7}
 & test & 1249 & 304 & 715 & 0 & 230  \\
\hline \multirow{3}{*}{Review} & train & 5485 & 2184 & 1222 & 210 & 2336 \\ \cline{2-7}
& val & 610 & 260 & 121 & 17 & 277 \\ \cline{2-7}
& test & 1446 & 496 & 455 & 60 & 634  \\
\hline 
\end{tabular} 
\end{center}
\label{tab:dataset}

\textit{\#pos} means the number of documents with positive sentiment for each target. If one document contains positive sentiment towards two targets, it will be counted twice. \textit{\#absent} counts the number of documents without any attitude towards any target. 
\end{table}

\vspace{5cm}

\subsection{Metrics}
For our problem, each data set has multiple targets, and each target can be classified into one of the outcomes: \textit{absent} (do not exist), \textit{neutral}, \textit{positive}, and \textit{negative}. If we treat each outcome of each target as one \textit{category}, we can adopt common metrics for multi-class classification. Since most targets do not appear in most instances, we have a highly skewed class distribution, where measures like accuracy are not good choices~\cite{chawla2005data}.


Apart from \textit{precision}, \textit{recall} and \textit{AUC}, we also use the \textit{macro-average F-measure}~\cite{yang1999re}. Let $\rho_i$ and $\pi_i$ be recall and precision respectively for a particular category $i$, $\rho_i = \frac{\textrm{TP}_i}{\textrm{TP}_i+\textrm{FN}_i}$, $\pi_i  = \frac{\textrm{TP}_i }{\textrm{TP}_i +\textrm{FP}_i }$, 
where $\textrm{TP}_i, \textrm{FP}_i, \textrm{FN}_i$ are the number of true positive, false positive, and false negative for category $i$. Given $\rho_i$ and $\pi_i$, F-score of category $i$ is computed as $F_i = \frac{2\pi_i\rho_i}{\pi_i+\rho_i}$. The macro-average F-score is obtained by taking the average over all categories. 
The final precision and recall are also averaged over individual categories. There is another micro-averaged F-measure, which is equivalent to accuracy. Therefore, we do not include it. 


\subsection{Baselines}
We compare baseline methods from two large categories: conventional methods and alternative deep learning methods. 

Each baseline method has various configurations, based on whether: (1) it trains \textit{a single model} or \textit{two separate models} for the $\vec{TD}$ and $\vec{PC}$ subtasks, and (2) it trains \textit{one universal model} for \textit{all} targets or \textit{separated models} for \textit{different} targets. To distinguish different configurations, we append \textit{-sgl} when using a single model for the two subtasks, and \textit{-sep} when using separate models for each subtask. 
Taking SVM as an example, SVM-sgl directly classify targets into four classes: \textit{absent}, \textit{neutral}, \textit{positive}, and \textit{negative}. In contrast, SVM-sep first classifies each target into two classes: \textit{absent}, \textit{present}, and use a second model to classify polarity: \textit{neutral}, \textit{positive}, and \textit{negative}. Moreover, we append \textit{-ind} when individual targets are trained on separate models, or \textit{-all} when one model is trained for all targets.

\subsubsection{Conventional baselines}
\textbf{SVM+features}. SVM using a set of hand-crafted features has achieved the state-of-the-art performance in stance classification of SemEval 2016  task~\cite{mohammad2016stance}, online debates~\cite{hasan2013stance}, 
and aspect-based sentiment analysis~\cite{tang2016aspect}. SVM has also demonstrated superior performance in document-level sentiment analysis compared with conditional random field methods~\cite{wang2012baselines}. Therefore we include all features from these methods that are general across domains, and use a linear kernel SVM implemented by LIBSVM~\cite{chang2011libsvm} for classification. We list the set of features:

\textit{Document info}: basic counting features of a document, including the number of characters, the number of words, the average words per document and the average word length.

\textit{N-grams}: word unigrams, bigrams, and trigrams. We insert symbols that represent the start and end of a document to capture cue words~\cite{walker2012your}.

\textit{Sentiment}: the number of positive and negative words counted from the NRC Emotion Lexicon~\cite{mohammad2010emotions}, Hu and Liu
Lexicon~\cite{hu2004mining}, and the MPQA Subjectivity Lexicon~\cite{wilson2005recognizing}.

\textit{Target}: presence of the target phrase in the text. Furthermore, if the target is present, we generate a set of target dependent features according to~\cite{jiang2011target}. To get a sense of these features, for the target \textit{iPhone} in text \textit{I love iPhone}, a feature \textit{love\_arg} will be generated.

\textit{POS}: the number of occurrences of each part-of-speech tag (POS).

\textit{Syntactic dependency}: a set of triples obtained by Stanford dependency parser~\cite{de2006generating}. More specifically, the triple is of the form $(rel, w_i, w_j)$, where $rel$ represents the grammatical relation between word $w_i$ and $w_j$, e.g., \textit{is subject of}.

\textit{Generalized dependency}: the first word of the dependency triple is ``backed off'' to its part-of-speech tag~\cite{walker2012stance}. Additionally, words that appear in sentiment lexicons are replaced by positive or negative polarity equivalents~\cite{walker2012stance}.

\textit{Embedding}: the element-wise averages of the word vectors for all the words in a document. We use three types of word embeddings. Two of them are from studies on target-dependent sentiment classification~\cite{vo2015target,zhang2016gated}, which are the skip-gram embeddings of Mikilov et al. \cite{mikolov2013distributed} and the sentiment-driven embeddings of Tang et al.~\cite{tang2014learning}. The first type of embedding is trained on 5 million unlabeled tweets that contain emoticons, which guarantees that more sentiment related tweets are included. The second type of embedding is of 50 dimensions, which is publicly available\footnote{\url{http://ir.hit.edu.cn/~dytang/}}. The third type of embedding is also 50-dimensional, released by Collobert et al.~\cite{collobert2011natural} and trained on English Wikipedia\footnote{\url{http://ronan.collobert.com/senna/}}.

\textit{Word cluster}: the number of occurrences of word clusters for all words in text. We perform K-means clustering on the word vectors. 

Apart from two standard SVM model configurations, \textit{SVM-sep-ind} and \textit{SVM-sgl-ind}, we also compare with a hybrid model \textit{SVM-cmb-ind}, whose prediction is \textit{absent} if \textit{SVM-sep-ind} says so, and otherwise it follows the decisions of \textit{SVM-sgl-ind}.\footnote{\textit{SVM-sgl-all} and \textit{SVM-sep-all} have performance degeneration due to the interference of different targets. We do not include their results for simplicity.}

\subsubsection{Deep Learning Baselines}
\textbf{BiLSTM}, \textbf{MultiBiLSTM} and \textbf{Memnet}. We also compare to the bidirectional LSTM (\textbf{BiLSTM}) model, the state-of-the-art in target-dependent sentiment classification~\cite{zhang2016gated}. Their variant of BiLSTM model assumes that the given target always appears exactly once, and can be tagged in text by starting and ending offsets. When such assumption fails, their model is equivalent to standard BiLSTM. We include the standard multi-layered bidirectional LSTM (\textbf{MultiBiLSTM})~\cite{irsoy2014deep} as an extension. Recently, Tang et al.~\cite{tang2016aspect} applied memory networks (\textbf{Memnet}) to multi-aspect sentiment analysis. Their results show memory network performs comparably with feature based SVM and outperforms all LSTM-related methods in their tasks.

\textbf{CNN} and \textbf{ParaVec}. We include related deep learning techniques beyond the sentiment analysis domain, such as the convolutional neural networks (\textbf{CNN})~\cite{kim2014convolutional} and  \textbf{ParaVec}~\cite{le2014distributed}. \textbf{ParaVec} requires a huge amount of training data to reach decent performance.  We enhance the performance of the ParaVec model by training over the merged training set of all data sets, plus the 5 million unlabeled tweets mentioned above.

Parser-dependent deep learning methods have also been applied to sentiment analysis~\cite{socher2012semantic,socher2013recursive,hermann2013role}. These models are limited in our attitude identification problem for two reasons. First, they often work well with phrase-level sentiment labels, but only  document-level sentiment labels are provided in our problems. Second, their parsers do not extend to user generated content, such as Tweets and Debates~\cite{gimpel2011part}. Our preliminary results show these methods work poorly on our problems and we do no include their results for simplicity.

For all deep learning methods, we report their \textit{-sep-all} and \textit{-sgl-all} version. Unlike SVM, deep methods perform quite well when using a single model for all targets, by casting the problem as a multi-task multi-class classification. Though not scalable, for the strongest baselines (BiLSTM and MultiBiLSTM), we in addition train a separate model for each target. Since \textit{-sep-ind} works better than \textit{-sgl-ind}, we only report the former one. 
The variants of memory networks are detailed below. 


\subsection{Variants of AttNet}
To assign the credit of key components in our model, we construct a competing model \textit{AttNet-}. Unlike our proposed model, for \textit{AttNet-} the target-specific projection matrices $\vec{V}_{q}^{p}$ and $\vec{V}_{q}^{t}$ are replaced by the identity matrix and are fixed during training. Thus the \textit{AttNet-} model interleave the target detection and polarity classification subtasks, but do not consider the interactions among targets. We refer our proposed model as \textit{AttNet}, which allows the projection matrices to be learned during training, and thus word semantics could vary for targets.

For \textit{AttNet-}, we report two settings in our experiments: \textit{AttNet-ind} and \textit{AttNet-all}. The former makes all targets share the same embedding, while the latter separates the embedding space completely for each target, i.e., targets are trained on separate models.

\begin{table}[h!]
\caption{Hyper-parameters for our method AttNet.}
\label{tab:exp_params}
\begin{center}
\small
\begin{tabular}{c|c|c|c} 
\hline 
Hyper-parameters & Tweets & Review & Debates \\
\hline 
L1 coeff & 1e-6 & 1e-4 & 1e-6 \\
\hline 
L2 coeff & 1e-4 & 1e-8 & 1e-8 \\
\hline 
init learning rate & 0.05 & 0.01 & 0.005 \\
\hline 
\#layers(target) & 4 & 4 & 3 \\
\hline 
\#layers(sentiment) & 4 & 8 & 6 \\
\hline 
prior attention $\lambda$ & 0.5 & 0.1 & 0.5 \\
\hline 
\end{tabular}

\flushleft The embedding size is set to 100 for all data sets. The sliding window size of the moving average function in Equation~\ref{equ:prior_attention} is set to 3.
\textit{\#layers(target)} is the number of memory layers  for target detection, and \textit{\#layers(sentiment)} is the number for sentiment classification. 
\textit{prior attention $\lambda$} is the weight for prior attention in Equation~\ref{equ:prior_attention}.
\end{center}
\end{table}

\begin{table*}[t]
\caption{Performance of competing methods: AttNet achieves top performance.}
\label{tab:exp_results}
\begin{center}
\setlength{\tabcolsep}{3pt}
\renewcommand{\arraystretch}{1.1}
\tiny
\begin{tabular}{c|c|c|c|c||c|c|c|c||c|c|c|c}  
\hline
 & \multicolumn{4}{c||}{Tweets} &  \multicolumn{4}{|c||}{Review}  &  \multicolumn{4}{|c}{Debates} \\
\hline
Method & F-score & AUC  & Precision  & Recall & F-score & AUC  & Precision  & Recall & F-score & AUC  & Precision  & Recall\\
\hline 
SVM-sep-ind & $59.93$ & $69.20$ & $68.70$ & $55.69$ & $38.43$ & $57.99$ & \boldmath$51.22$ & $36.83$ & $58.30$ & $72.10$ & $64.48$ & $57.81$ \\
SVM-sgl-ind & $57.44^{***}$ & $66.64^{***}$ & \boldmath$69.87$ & $52.45^{***}$ & $36.06^{**}$ & $56.84^{**}$ & $50.79$ & $34.07^{**}$ & $59.75$ & $72.25$ & $66.39$ & $57.67$ \\
SVM-cmb-ind & $57.09^{***}$ & $66.46^{***}$ & $69.84$ & $52.25^{***}$ & $35.71^{***}$ & $56.61^{***}$ & $50.73$ & $33.78^{***}$ & $59.86$ & $71.68$ & $66.28$ & $56.48$ \\
\hline 
ParaVec-sep-all & $53.17^{***}$ & $62.88^{***}$ & $56.75^{***}$ & $48.29^{***}$ & $34.02^{***}$ & $55.26^{***}$ & $38.04^{***}$ & $30.47^{***}$ & $56.32^{**}$ & $68.12^{***}$ & $59.09^{***}$ & $49.41^{***}$ \\
ParaVec-sgl-all & $54.15^{***}$ & $63.41^{***}$ & $57.52^{***}$ & $48.76^{***}$ & $34.26^{***}$ & $55.31^{***}$ & $38.26^{***}$ & $30.89^{***}$ & $55.35^{***}$ & $67.48^{***}$ & $59.46^{***}$ & $49.82^{***}$ \\
\hline 
CNN-sep-all & $58.05^{*}$ & $70.10$ & $62.43^{***}$ & $56.19$ & $37.15^{*}$ & $57.55$ & $43.73^{***}$ & $33.24^{**}$ & $57.38$ & $70.70^{**}$ & $61.81^{**}$ & $52.94^{***}$ \\
CNN-sgl-all & $58.69$ & $70.71^{*}$ & $61.83^{***}$ & $56.64$ & $35.45^{***}$ & $56.29^{*}$ & $44.65^{***}$ & $32.83^{**}$ & $56.23^{**}$ & $69.92^{**}$ & $60.75^{**}$ & $52.29^{***}$ \\
\hline 
BiLSTM-sep-all & $61.16$ & $71.26^{**}$ & $63.45^{*}$ & $59.87^{***}$ & $40.78^{*}$ & $61.01^{***}$ & $42.54^{***}$ & $39.01^{**}$ & $59.83^{*}$ & $71.91$ & $65.94^{*}$ & $57.65$ \\
BiLSTM-sgl-all & $60.86$ & $71.02^{**}$ & $62.58^{*}$ & $59.61^{***}$ & $39.68$ & $60.84^{**}$ & $41.88^{***}$ & $38.81^{*}$ & $58.66$ & $72.01$ & $64.87$ & $57.89$ \\
BiLSTM-sep-ind & $59.49$ & $71.92^{**}$ & $61.44^{***}$ & $57.86$ & $40.42$ & $62.25^{***}$ & $42.68^{***}$ & $39.78^{**}$ & $58.75$ & $72.83$ & $64.73$ & $57.95$ \\
\hline 
MultiBiLSTM-sep-all & $60.53$ & $71.51^{**}$ & $64.81^{*}$ & $57.76$ & $40.47$ & $60.71^{**}$ & $44.89^{***}$ & $37.67^{*}$ & $59.24^{*}$ & $72.24$ & $64.75$ & $58.43$ \\
MultiBiLSTM-sgl-all & $60.59$ & $71.32^{**}$ & $64.27^{*}$ & $57.97$ & $39.38$ & $59.68^{*}$ & $43.22^{***}$ & $37.92^{*}$ & $58.98$ & $71.18$ & $63.46$ & $57.26$ \\
MultiBiLSTM-sep-ind & $59.71$ & $71.16^{**}$ & $63.72^{*}$ & $57.94$ & $40.81$ & $61.27^{**}$ & $44.76^{***}$ & $38.02^{*}$ & $58.36$ & $72.93$ & $64.15$ & $57.14$ \\
\hline 
Memnet-sep-all & $59.44$ & $71.68^{**}$ & $63.22^{*}$ & $59.80^{***}$ & $41.75^{**}$ & $61.82^{***}$ & $45.61^{***}$ & $39.25^{**}$ & $60.42^{**}$ & $73.84^{*}$ & $65.37$ & $58.92^{*}$ \\
Memnet-sgl-all & $60.69$ & $71.80^{**}$ & $63.48^{*}$ & $59.97^{***}$ & $41.65^{**}$ & $61.53^{***}$ & $45.23^{***}$ & $39.13^{**}$ & $59.67^{*}$ & $73.31^{**}$ & $64.27$ & $58.83$ \\
\hhline{=|=|=|=|=#=|=|=|=#=|=|=|=}
\multicolumn{1}{l|}{\tiny Proposed methods}&&&&&&&&&&&&\\
AttNet-all & $63.42^{**}_{\diamond\diamond\diamond}$ & $72.94^{***}_{\diamond}$ & $68.78_{\diamond\diamond\diamond}$ & $60.57^{***}$ & $43.91^{***}_{\diamond\diamond}$ & $63.18^{***}_{\diamond\diamond}$ & $47.89^{**}_{\diamond\diamond}$ & $42.77^{***}_{\diamond\diamond}$ & $64.23^{***}_{\diamond\diamond\diamond}$ & $76.13^{***}_{\diamond\diamond}$ & $67.19^{**}_{\diamond\diamond}$ & $62.17^{**}_{\diamond\diamond\diamond}$ \\
AttNet-ind & $63.09^{**}_{\diamond\diamond\diamond}$ & $72.73^{***}$ & $68.33_{\diamond\diamond\diamond}$ & $59.68^{**}$ & $44.15^{***}_{\diamond\diamond}$ & $63.53^{***}_{\diamond\diamond}$ & $50.02_{\diamond\diamond\diamond}$ & $40.58^{**}$ & $65.01^{***}_{\diamond\diamond\diamond}$ & $76.35^{***}_{\diamond\diamond}$ & $70.08^{***}_{\diamond\diamond\diamond}$ & $60.70^{*}_{\diamond}$ \\
AttNet & \boldmath$64.62^{***}_{\scriptscriptstyle\triangledown}$ & \boldmath$74.76^{***}_{\scriptscriptstyle\triangledown\triangledown}$ & $68.40$ & \boldmath$62.09^{***}_{\triangledown\triangledown}$ & \boldmath$45.93^{***}_{\triangledown}$ & \boldmath$65.58^{***}_{\triangledown\triangledown}$ & $50.34$ & \boldmath$44.95^{***}_{\triangledown\triangledown}$ & \boldmath$67.68^{***}_{\triangledown\triangledown}$ & \boldmath$78.48^{***}_{\triangledown\triangledown\triangledown}$ & \boldmath$74.55^{***}_{\triangledown\triangledown\triangledown}$ & \boldmath$66.31^{***}_{\triangledown\triangledown}$ \\
\hline
\end{tabular}
\small
\flushleft*(**,***) indicate that one method is statistically significantly better or worse than \textit{SVM-sep-ind} (which is in general the best configuration among all SVM models) according to t-test~\cite{yang1999re} at the significance level of 0.05(0.01,0.001). $\scriptscriptstyle\diamond\diamond(\diamond\diamond\diamond)$ indicate \textit{AttNet-} outperforms the better one between \textit{Memnet-sep-all} and \textit{Memnet-sgl-all} at the significance level of 0.01(0.001). $\scriptscriptstyle\triangledown\triangledown(\triangledown\triangledown\triangledown)$ indicate \textit{AttNet} outperforms the better one between \textit{AttNet-all} and \textit{AttNet-ind} at the significance level of 0.01(0.001).
\end{center}
\renewcommand{\arraystretch}{1}
\end{table*}

\subsection{Training Details}
All hyper-parameters are tuned to obtain the best performance of F-score  on validation set. The candidate embedding size set is $\{50,$ $100,$ $200,$ $300\}$ for LSTM-related methods, SVM and CNN. The candidate number of clusters for K-means is $\{50,$ $100,$ $150\}$.  The candidate relaxing parameter C for SVM model is $\{2^7, 2^6, ..., 2^{-3}\}$. The CNN model has three convolutional filter sizes and their filter size candidates are $\{\{1,2,3\}$, $\{2,3,4\}$, $\{3,4,5\}$, $\{2,4,6\}\}$, and  the candidate number of filters is $\{50,$ $100,$ $200,$ $300\}$. For ParaVec, we experiment with both skip-gram model or bag-of-words model, and select the hidden layer size from $\{2^6,$ $2^7,$ $...,$ $2^{10}\}$. 

We explored three weight initialization methods of word embeddings for LSTM-related and CNN baselines: (1) sampling weights from a zero-mean Gaussian with 0.1 standard deviation; (2) initializing from the pre-trained embedding matrix, and (3) using a fixed pre-trained embedding matrix. 

Memory network models, including our model, are initialized by sampling weights from a zero-mean Gaussian with unit standard deviation. The candidate number of memory layers is $\{2,$ $3,...,$ $9\}$. The  prior attention parameter $\lambda$ of our model is selected from $\{0,0.1,0.5,0.9,1\}$.  The capacity of memory, which has limited impact on the performance, is restricted to 100 words without further tuning. A null symbol was used to pad all documents to this fixed size. To reduce the model complexity, the projection matrices are initialized in the way that each column is a one-hot vector.


Deep learning models are optimized by Adam~\cite{kingma2015adam}.  The initial learning rate is selected from $\{0.1, 0.05, 0.01, 0.005, 0.001\}$, and L1-coefficient and L2-coefficient of regularizers are selected from $\{10^{-2},$ $10^{-4},$ $...,$ $10^{-10}\}$. The hyper-parameters of our model \textit{AttNet+} for different data sets are listed in Table~\ref{tab:exp_params}.



%% file: tex/Experiment_result.tex
\section{Experiment results}
\label{sec:exp}

\subsection{Overall Performance}
The overall performance of all competing methods over data sets are shown in Table~\ref{tab:exp_results}\footnote{The performance of all methods on the Review data set is lower than the other two because Review data set handles three polarities while the others only need to handle two polarities as shown in Table~\ref{tab:dataset}.}. Evaluating with F-score and AUC, we make the following observations. Our method \textit{AttNet} outperforms all competing methods significantly. This empirically confirms that interleaving target detection and polarity classification subtasks combined with target-specific representations could benefit attitude identification. 

The variants of our model, \textit{AttNet-all} and \textit{AttNet-ind}, have already gained significant improvement over the strongest baselines on all data sets. More importantly, the two methods significantly outperform the \textit{Memnet-sep-all} and \textit{Memnet-sep-all} baselines, which do not interleave the subtasks. Such empirical findings cast light on that interleaving the subtasks indeed improves the attitude identification performance. In contrast, separating the two subtasks of attitude identification could lead to performance degeneration. 

Our model \textit{AttNet} also outperforms its variants, \textit{AttNet-all} and  \textit{AttNet-ind}, on all data sets. The performance advantage of \textit{AttNet} comes from the adoption of target-specific projection matrices in representation learning, since these matrices are the only differences between \textit{AttNet} and \textit{AttNet-}.  Even though the improvement from adopting target-specific projection matrices is not as marked as from the techniques of interleaving the subtasks, the improvement is still significant. This result confirms that attitude identification could benefit from the learned representations that share the same semantics for many targets but vary for some targets.

By examining the \textit{precision} and \textit{recall} results, we find that the superior performance of our model is mainly from the significant improvement of \textit{recall}, though both \textit{precision} and \textit{recall} are improved significantly on the Debates data set.

\subsection{Performance on Subtasks}
We have established that our models outperform competing methods on all data sets. In order to further assign the credits of the improvement of our methods, we evaluate our models on the two subtasks: target detection and polarity classification, with results given in Table~\ref{tab:exp_task1_results} and \ref{tab:exp_task2_results} respectively. Since different configurations of the same method work similarly, we only present the results where separate models are trained for each task. 
It can be seen from Table~\ref{tab:exp_task1_results} that the target detection task is relatively easy, as all methods can achieve quite high scores. This also means that it is hard to improve any further on this task. In terms of precision and recall, SVM performs quite well on the precision metric, especially for the Review data set. While most deep learning methods focus more on enhancing recall. When considering both precision and recall, most deep learning methods are still better, as the F-score shows.

\begin{table}[h!]
\caption{Performance on target detection for \textit{-sep} models.}
\label{tab:exp_task1_results}
\begin{center}
\setlength{\tabcolsep}{2.5pt}
\scriptsize
\begin{tabular}{c|c|c|c} 
\hline 
 & Tweets & Review & Debates \\
\hline 
\multirow{2}{*}{SVM}   & $79.74$ & $67.84$ & $84.59$ \\
 & \tiny \boldmath$89.12$\unboldmath,$75.00$ & \tiny\boldmath$84.13$\unboldmath,$63.52$ & \tiny$91.47$,$81.00$ \\
\hline 
\multirow{2}{*}{ParaVec} & $75.63^{**}$ & $63.74^{***}$ & $76.16^{***}$ \\
 & \tiny$82.67^{***}$,$71.76^{***}$ & \tiny$67.41^{***}$,$57.03^{***}$ & \tiny$80.32^{***}$,$71.80^{***}$ \\
\hline 
\multirow{2}{*}{CNN} & $80.62$ & $65.34^{**}$ & $80.21^{**}$ \\
 & \tiny$87.57^{**}$,$77.41^{*}$ & \tiny$73.23^{***}$,$59.16^{**}$ & \tiny\boldmath$93.26^{*}$\unboldmath,$73.48^{**}$ \\
\hline 
\multirow{2}{*}{BiLSTM} & $81.50^{**}$ & $70.41^{**}$ & $85.05$ \\
 & \tiny$86.48^{*}$,$81.68^{**}$ & \tiny$76.39^{***}$,$69.11^{***}$ & \tiny$92.03$,$82.05$ \\
\hline 
\multirow{2}{*}{MultiBiLSTM} & $81.53^{**}$ & $69.26^{**}$ & $85.33$ \\
 & \tiny$87.69$,$78.82^{**}$ & \tiny$75.38^{***}$,$68.40^{***}$ & \tiny$92.82$,$83.77^{*}$ \\
\hline 
\multirow{2}{*}{Memnet} & $81.29^{**}$ & $70.71^{**}$ & $86.29^{*}$ \\
 & \tiny$87.82$,$79.36^{**}$ & \tiny$75.52^{***}$,$68.59^{***}$ & \tiny$92.31$,$83.26$ \\
\hline 
\multirow{2}{*}{AttNet-all} & $82.58^{**}$ & $71.84^{***}$ & $89.02^{***}_{\diamond\diamond}$ \\
 & \tiny$88.78$,$79.05^{**}$ & \tiny$75.85^{***}$,$69.99^{***}$ & \tiny$92.24$,$86.47^{***}_{\diamond\diamond}$ \\
\hline 
\multirow{2}{*}{AttNet-ind} & $82.74^{**}_{\diamond}$ & $71.95^{***}$ & $88.89^{**}_{\diamond\diamond}$ \\
& \tiny$88.29$, $79.82^{**}$ & \tiny$82.67_{\diamond\diamond\diamond}$, $66.19^{**}$ & \tiny$92.35$, $82.22$ \\
\hline 
\multirow{2}{*}{AttNet} & \boldmath$84.89^{***}_{\triangledown\triangledown}$ & \boldmath$72.59^{***}$ &\boldmath$89.35^{***}$ \\
 & \tiny$88.76$,\boldmath$82.24^{***}_{\triangledown\triangledown}$ & \tiny$76.38^{***}$,\boldmath$71.09^{***}_{\triangledown}$ & \tiny$92.05$,\boldmath$87.12^{***}$ \\
\hline 
\end{tabular}
\flushleft The first row of each method shows F-score, followed by precision and recall on the second row.
\end{center}
\end{table}

The second task is only evaluated on documents with ground-truth sentiments towards particular targets, with F-scores averaged over all targets and three sentiment classes: \textit{positive}, \textit{negative}, and \textit{neutral}. We make several notes for this evaluation. (1) In order to achieve a high score in the second task, it is still important to classify correctly the presence of a target. (2) In general the scores for all methods in the second task are low, due to that the classifier might predict a target as \textit{absent}, even though the ground-truth class can only be drawn from three sentiment classes. (3) It is possible for a method to outperform SVM on both tasks, while still obtain close results when two tasks are evaluated together. This results from our method of evaluation on the second task, where a document is included only when it expresses sentiment towards a particular target.

Based on the results from Table~\ref{tab:exp_task2_results}, we can see that the percentage of improvement over SVM is much higher than that of the first task. Intuitively, the sentiment task requires better modeling of the non-linear interaction between the target and the context, while for the target detection task, presence of certain signal words might be enough. 

\begin{table}[h!]
\caption{Performance on polarity classification for \textit{-sep} models.}
\label{tab:exp_task2_results}
\setlength{\tabcolsep}{2.5pt}
\scriptsize
\begin{center}
\begin{tabular}{c|c|c|c} 
\hline 
 & Tweets & Review & Debates \\
\hline 
\multirow{2}{*}{SVM} & $44.45$ & $21.37$ & $42.52$  \\
 & \tiny$64.50$,$34.28$ &\tiny $53.25$,$14.66$ &\tiny $57.37$,$38.29$ \\
\hline 
\multirow{2}{*}{ParaVec} & $39.20^{***}$ & $17.69^{***}$ & $30.59^{***}$  \\
 & \tiny$56.91^{**}$,$26.46^{***}$ & \tiny$31.05^{***}$,$9.25^{***}$ & \tiny$56.15$,$20.88^{***}$ \\
\hline 
\multirow{2}{*}{CNN} & $42.95$ & $19.42^{*}$ & $35.15^{***}$  \\
 & \tiny$58.20^{**}$,$35.71$ & \tiny$39.91^{**}$,$11.34^{**}$ &\tiny $49.82^{**}$,$25.40^{***}$ \\
\hline 
\multirow{2}{*}{BiLSTM} & $46.18^{**}$ & $25.25^{**}$ & $42.69$  \\
 & \tiny$61.35^{*}$,$41.79^{***}$ & \tiny$41.06^{***}$,$19.40^{**}$ & \tiny$54.58$,$35.59$ \\
\hline 
\multirow{2}{*}{MultiBiLSTM} & $46.26^{**}$ & $24.06^{**}$ & $41.87$  \\
 & \tiny$60.56^{**}$,$39.36^{**}$ & \tiny$47.15^{**}$,$17.52^{**}$ & \tiny$49.69^{*}$,$37.26$ \\
\hline 
\multirow{2}{*}{Memnet} & $47.81^{**}$ & $25.47^{**}$ & $44.61$  \\
 & \tiny$61.20^{**}$,$40.52^{**}$ & \tiny$46.34^{**}$,$19.81^{**}$ & \tiny$54.40^{*}$,$38.14$ \\
\hline 
\multirow{2}{*}{AttNet-all} & $50.91^{***}_{\diamond\diamond}$ & $32.43^{***}_{\diamond\diamond\diamond}$ & $50.46^{***}_{\diamond\diamond\diamond}$  \\
 & \boldmath\tiny$66.39_{_{\diamond\diamond\diamond}}$\unboldmath,$42.00^{***}_{\diamond}$ & \tiny$55.92_{\diamond\diamond\diamond}$,$24.43^{***}_{\diamond\diamond}$ & \tiny$60.72^{*}_{\diamond\diamond\diamond}$,$44.02^{**}_{\diamond\diamond}$ \\
\hline 
\multirow{2}{*}{AttNet-ind} & $49.16^{**}_{\diamond}$ & $32.79^{***}_{\diamond\diamond\diamond}$ & $51.88^{***}_{\diamond\diamond\diamond}$ \\
 & \tiny$64.01_{\diamond\diamond}$, $41.74^{***}$ & \tiny$56.15^{**}_{\diamond\diamond}$, $23.17^{***}_{\diamond\diamond}$ & \tiny$62.70^{**}_{\diamond\diamond\diamond}$, $40.52$ \\
\hline 
\multirow{2}{*}{AttNet} & \boldmath$52.23^{***}_{\triangledown\triangledown}$ & \boldmath$35.34^{***}_{\triangledown\triangledown}$ & \boldmath$55.93^{***}_{\triangledown\triangledown\triangledown}$  \\
 & \tiny$65.54$,\boldmath$44.57^{***}_{\triangledown\triangledown}$ & \tiny\boldmath$59.99^{***}_{\triangledown\triangledown}$,$27.32^{***}_{\triangledown\triangledown}$ & \tiny\boldmath$71.53^{**}_{\triangledown\triangledown}$,$50.25^{***}_{\triangledown\triangledown}$ \\
\hline 
\end{tabular}

\flushleft The first row of each method shows F-score, followed by precision and recall on the second row.
\end{center}
\end{table}

\subsection{Training Time Analysis}
In order to measure the training speed of each model, we train all deep learning methods on a server with a single TITAN X GPU. For SVM, it is trained on the same server with a 2.40 GHz CPU and 120 G RAM. All methods are trained sequentially without parallelization. 

%

SVM can finish training in less than one hour, but its required training time increase linearly as the number of targets increases.

For all deep learning methods, the number of epochs required for training is in general very close, which is around 20 epochs averaged over all data sets.

Comparing the training time per epoch, ParaVec and CNN are much faster than other methods (less than 5 seconds / epoch). Despite the training efficiency, their effectiveness is a problem. When all targets share a single model, LSTM has a speed of 200 seconds/epoch, while standard memory networks have a speed of 150 seconds/epoch. Memory networks in many tasks, e.g., language modeling, are much faster than LSTM, due to the expensive recursive operation of LSTM. However in our problem setting, each target has to be forwarded one by one for every document, lowering the efficiency of memory networks. When individual targets are trained on separate LSTMs, LSTMs require far more training time (1000 seconds/epoch).

\textit{AttNet} consumes 200 seconds per epoch. Comparing to standard memory networks, \textit{AttNet} produces some additional overhead by introducing the interaction between subtasks, and by adding a projection matrix. But this overhead is small.


The efficiency of deep learning methods could be improved by parallelization. Since there are already many work on this topic, which could increase the speed without sacrificing effectiveness, we do not go further into this direction. \\

\noindent \textbf{Summary}: empirical experiments demonstrated that the proposed deep memory networks, AttNet and its variants outperforms conventional supervised learning methods. This is promising but perhaps not surprising given the success of deep learning in general. It is encouraging to notice that AttNet also improves the state-of-the-art deep learning architectures. This improvement is statistically significant, and can be observed for both subtasks and for attitude identification as a whole. The improvement of effectiveness does not compromise learning efficiency. 

\subsection{Visualization of attention}
In order to better understand the behavior of our models, we compare the attention weights given by our model AttNets and the competing method Memnet.

\begin{figure}[h]
\centering
\includegraphics[width=0.45\textwidth]{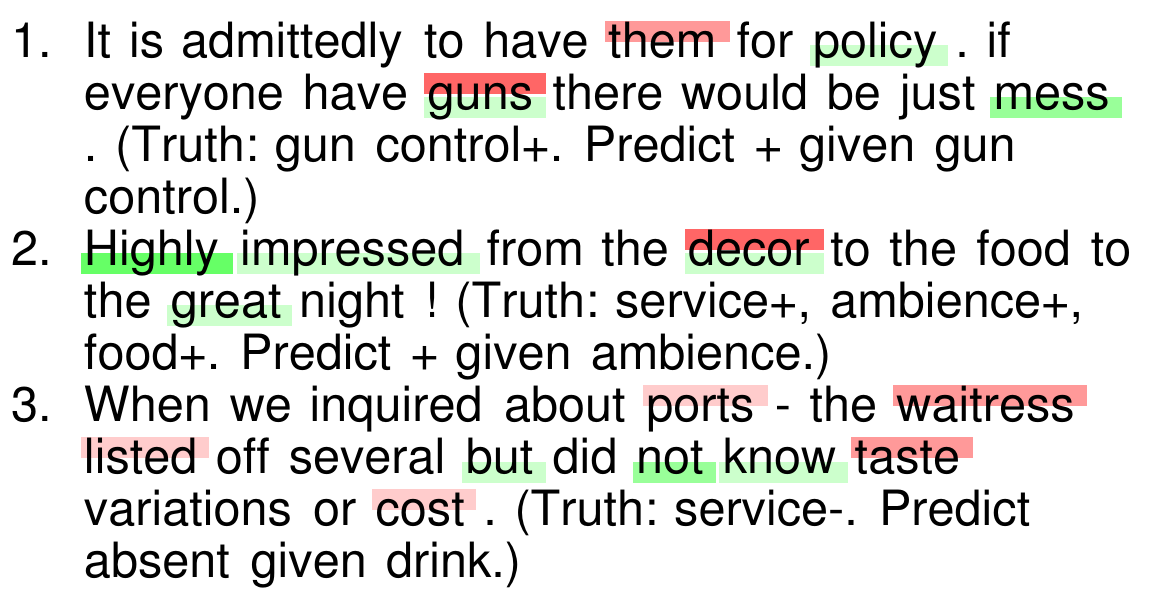}

(a) Attention given by AttNets.

\includegraphics[width=0.45\textwidth]{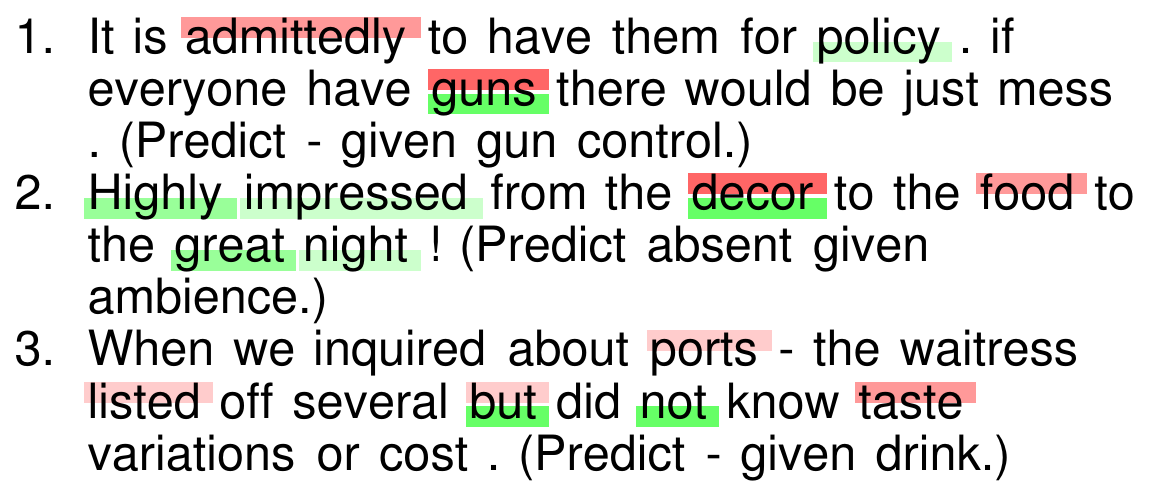}

(b) Attention given by Memnet.

\caption{Visualization of learned attention. Red patches highlighting the top half of the text indicate model's attention weight in the target detection task, while green ones highlighting the bottom half show the polarity classification task. Darker colors indicate higher attentions. \textit{Truth: service+} means that the ground-truth sentiment towards \textit{service} is positive, while \textit{Predict + given ambience} gives the predicted positive sentiment given the query target \textit{ambience}.} 
\label{fig:vis_attention}
\end{figure}

Figure~\ref{fig:vis_attention} (a) and (b) list some examples of word attentions generated by different models for the same set of sentences in the test set. In the first sentence, both \textit{them} and \textit{guns} are found as targets by AttNets, while words like \textit{mess} and \textit{policy} are found as sentiment words. Though Memnet correctly identifies the existence of the attitude towards gun control, it fails to find important words to classify the polarity of sentiment. This suggests the importance of interleaving the two tasks -- successfully identifying mentioned targets could offer clues about the finding of sentiment words for the second task.

The second sentence is from a review of a restaurant, when \textit{ambience} is used as the query target. We can see that the target detection module of AttNets captures the word \textit{decor}, which signals the presence of the target \textit{ambience}. The polarity classification module then focuses on extracting sentiment words associated with the target. However for the baseline Memnet, it captures both \textit{decor} and \textit{food} in the first task, mistakenly considering all sentiments are only describing food other than the ambience. Consequently, it judges that there is no attitude towards ambience. This example shows us the benefit of using the projection matrices to consider the interaction and distinction between targets. Otherwise the model might easily be confused by to which entity the sentiments are expressed.

From the third sentence, we can see how our model AttNets determines that the query target \textit{drink} does not exist. The first module highlights words like ports (wine name), waitress, and the second module extracts negative sentiments \textit{but not know}, which is usually used to describe people, rather than drink. Memnet almost has the same attention distribution as AttNets, but still fails to produce the correct prediction. Similar to the second case, projection matrices are important for models to figure out the common phrases used to describe different set of entities.

%% file: tex/Conclusion.tex
\section{Conclusion}
\label{sec:conclusions}
Attitude identification, a key problem of morden natural language processing, is concerned with detecting one or more target entities from text and then classifying the sentiment polarity towards them. This problem is conventionally approached by separately solving the two subtasks and usually separately treating each target, which fails to leverage the interplay between the two subtasks and the interaction among the target entities. Our study demonstrates that modeling these interactions in a carefully designed, end-to-end deep memory network significantly improves the accuracy of the two subtasks, target detection and polarity classification, and attutide indentification as a whole. Empirical experiments proves that this novel model outperforms models that do not consider the interactions between the two subtasks or among the targets, including conventional methods and the state-of-the-art deep learning models. 

This work opens the exploration of interactions among subtasks and among contexts (in our case, targets) for sentiment analysis using an end-to-end deep learning architecture. Such an approach can be easily extended to handle other related problems in this domain, such as opinion summarization, multi-aspect sentiment analysis, and emotion classification. Designing specific network architecture to model deeper dependencies among targets is another intriguing future direction.    

\subsection*{Acknowledgment} This work is partially supported by the National Science Foundation under grant numbers IIS-1054199 and SES-1131500.